\documentclass[journal]{new-aiaa}
\usepackage[utf8]{inputenc}
\usepackage{graphicx}
\usepackage{amsmath}
\usepackage{siunitx}
\usepackage{longtable,tabularx}
\usepackage[capitalise]{cleveref}
\usepackage{physics}
\usepackage{booktabs}
\usepackage{multicol}
\usepackage{numprint}
\npthousandsep{,}\npthousandthpartsep{}\npdecimalsign{.}\npproductsign{\times}
\usepackage{dsfont}
\usepackage[dvipsnames]{xcolor}
\hypersetup{
linkcolor=blue!50!black,
citecolor=ForestGreen,
filecolor=Mulberry,
urlcolor=NavyBlue,
menucolor=BrickRed,
runcolor=Mulberry,
linkbordercolor=BrickRed,
citebordercolor=Green,
filebordercolor=Mulberry,
urlbordercolor=NavyBlue,
menubordercolor=BrickRed,
runbordercolor=Mulberry
}

\usepackage{tikz}
\renewcommand{\vec}{\mathbf}

\title{Bayesian Optimization of a Lightweight and Accurate Neural Network for Aerodynamic Performance Prediction}

\author{James M. Shihua}
\affil{Department of Mechanical and Aerospace Engineering, HKUST, Hong Kong}
\author{Paul Saves}
\affil{IRIT, Université Toulouse Capitole, CNRS, Toulouse INP, Université de Toulouse, Toulouse, France}

\author{Rhea P. Liem}
\affil{Department of Aeronautics, Imperial College London, United Kingdom}

\author{Joseph Morlier}
\affil{ICA, Université de Toulouse, ISAE-SUPAERO, MINES ALBI, UPS, INSA, CNRS, 3 rue Caroline Aigle, Toulouse, 31400, France}

\begin{document}

\maketitle

\begin{abstract}
\noindent
Ensuring high accuracy and efficiency of predictive models is paramount in the aerospace industry, particularly in the context of multidisciplinary design and optimization processes. 
These processes often require numerous evaluations of complex objective functions, which can be computationally expensive and time-consuming. 
To build efficient and accurate predictive models, we propose a new approach that leverages Bayesian Optimization (BO) to optimize the hyper-parameters of a lightweight and accurate Neural Network (NN) for aerodynamic performance prediction. 
To clearly describe the interplay between design variables, hierarchical and categorical kernels are used in the BO formulation.
We demonstrate the efficiency of our approach through two comprehensive case studies, where the optimized NN significantly outperforms baseline models and other publicly available NNs in terms of accuracy and parameter efficiency. 
For the drag coefficient prediction task, the Mean Absolute Percentage Error (MAPE) of our optimized model drops from 0.1433\% to 0.0163\%, which is nearly an order of magnitude improvement over the baseline model. 
Additionally, our model achieves a MAPE of 0.82\% on a benchmark aircraft self-noise prediction problem, significantly outperforming existing models (where their MAPE values are around 2 to 3\%) while requiring less computational resources.
The results highlight the potential of our framework to enhance the scalability and performance of NNs in large-scale MDO problems, offering a promising solution for the aerospace industry. 
\end{abstract}

\section{Introduction}
\label{sec:intro}
In the aerospace industry, Multidisciplinary Design and Optimization (MDO) processes commonly require a large number of objective function and constraint evaluations with the variation of design variables. 
These objective functions, which are typically solutions to complex differential equations or simulations, can be both time-consuming and costly to compute~\cite{martins2013multidisciplinary}. 
For certain MDO problems, a large number of evaluations may be required, making simulation within the optimization loop prohibitively expensive. 
To address this, surrogate models are employed to approximate these complex evaluations with simpler and more efficient representations, significantly speeding up the MDO process.

The Surrogate Modeling Toolbox (SMT)~\cite{SMT} is an open-source framework that facilitates the derivation and construction of surrogate models. 
A key feature of SMT is its support for various Gaussian process regression techniques and their interpolation counterpart, commonly known as Kriging. 
This includes ordinary Kriging~\cite{OK}, Kriging with partial least squares~\cite{KPLS}, gradient-enhanced Kriging~\cite{GEKPLS}, sparse Kriging~\cite{SGP}, and other variants. 
While Kriging models are adaptable to different data distributions and perform well with limited training samples, their computational cost increases rapidly with larger datasets.
As such, they are less ideal for scenarios where engineers attempt to achieve high model accuracy with extensive training data.

Regularized Minimal-energy Tensor-product Splines (RMTS)~\cite{RMTS} form a family of interpolation models that are computationally efficient for large datasets but are limited to low-dimensional problems. When dealing with large volumes of data, artificial Neural Networks (NN) provide a scalable alternative. Some successful examples of models that can leverage extensive datasets for training include the Gradient-Enhanced Neural Network (GENN)~\cite{GENN} and NeuralFoil~\cite{neuralfoil}.
In particular, GENN has been trained on high-fidelity data from Computational Fluid Dynamics (CFD) solvers, efficiently incorporating gradient information, whereas NeuralFoil has been applied to model mid-fidelity data from the viscous panel method~\cite{GENN,neuralfoil}.
Unlike Kriging models, where most Hyper-Parameters (HP) are optimized automatically, those for neural networks are not easy to train.
Neural networks are highly sensitive to HP, such as learning rate, number of layers, and batch size, which can significantly affect their performance and generalization capabilities~\cite{bengio2012practical}. 
Proper tuning of these HPs is crucial for achieving optimal model performance and avoiding issues such as overfitting or underfitting.
PyTorch-based tools like Ray Tune have become the industry standard for automatic and distributed hyperparameter tuning, particularly for large-scale hyperparameter optimization problems~\cite{bartz2023hyperparameter}. Additionally, other frameworks such as Optuna~\cite{optuna} and HyperOpt~\cite{hyperopt} provide standalone hyperparameter tuning capabilities while also integrating with Ray Tune. These frameworks typically employ algorithms such as grid search, genetic algorithms, and Bayesian optimization. Since evaluating model performance based on training accuracy is computationally expensive, Bayesian optimization is widely regarded as the most efficient and cost-effective approach~\cite{snoek2012practical}.

\begin{figure}[th!]
    \centering
    \includegraphics[width=\linewidth]{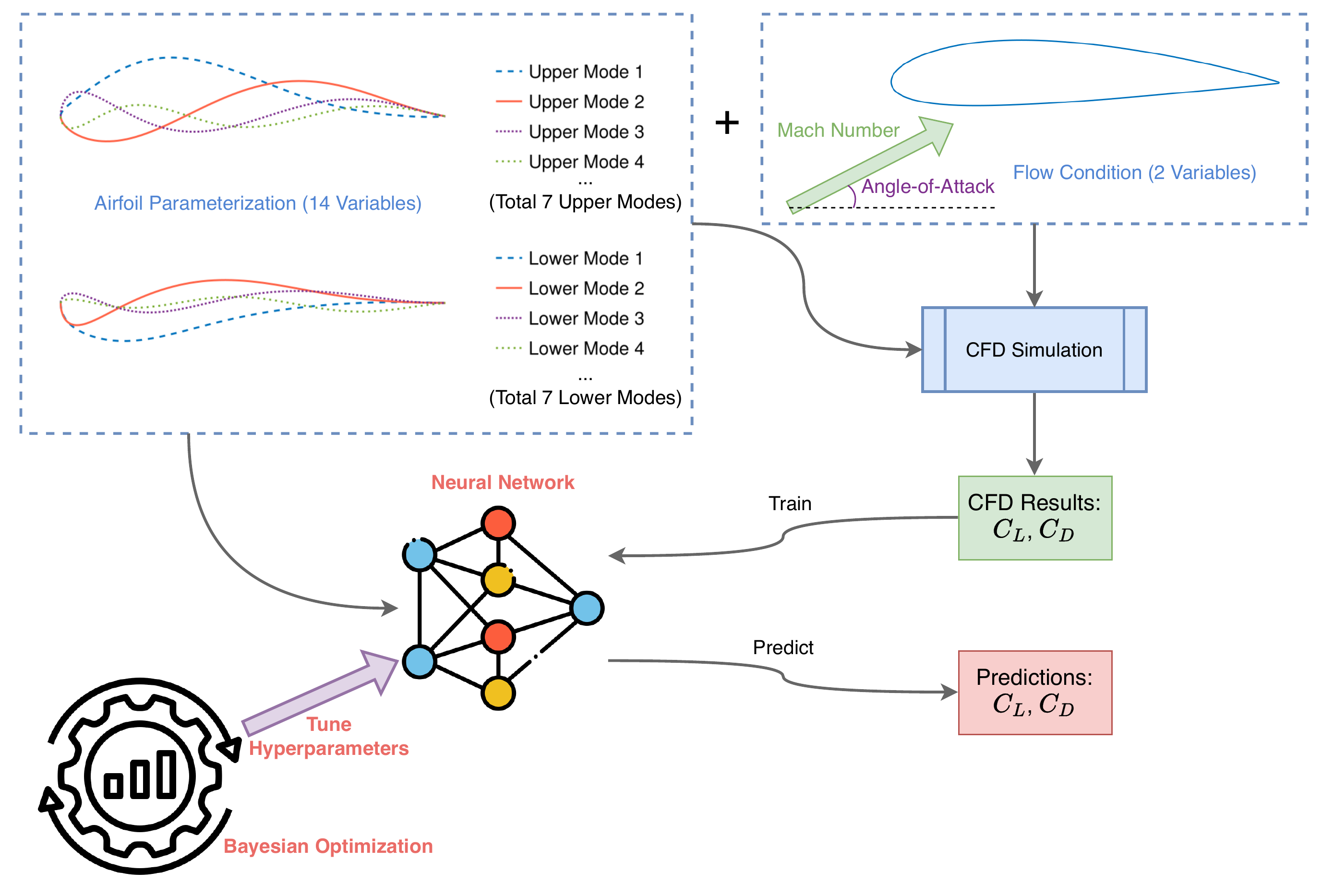}
    \caption{BO optimizes the hyper-parameters of the NN which predicts the aerodynamic performance from shape parameterization and flow condition.}
    \label{fig:ASO-BO}
\end{figure}

At present, most Bayesian optimization frameworks suffer from the following two key problems. 
First, they work predominantly on continuous variables, but many HPs are discontinuous. 
As an example, to find the suitable number of neurons in a layer, considering a sequence of numbers separated by an interval (\textit{e.g.}, $\{30,35,40,\ldots,80\}$) instead of a series of continuous integers is more practical as it can balance exploration and computational efficiency. In which case, a certain class of variables is needed for this variable, since simply setting it as an integer is not suitable.
Second, there are cases where the state of one variable is determined by another variable. 
As an example, when the user wants to optimize both the number of neurons per layer and the number of layers, the number of neurons in layer 3 is only necessary and relevant when there are at least three layers.
Considering these two problems, we propose to use SMT 2.0~\cite{SMT2} to tune HPs.
To address the first problem, the \texttt{DesignSpace} interface provides several special variable types: ordered (\texttt{OrdinalVariable}) and categorical (\texttt{CategoricalVariable}).
A discrete variable is ordinal if there is an order relation within the set of possible values and categorical if the sequence does not matter.
To tackle the second problem, SMT 2.0 also introduces several roles for each variable: meta, decreed, and neutral.
In particular, meta-variables control decreed variables according to the former's values, while neutral ones are independent.
Finally, the Efficient Global Optimization (EGO) algorithm leverages Bayesian optimization abilities for the above-mentioned variables with special types or roles.
\cref{fig:ASO-BO} graphically shows the workflow: BO optimizes the hyper-parameters of the NN which predicts the aerodynamic performance from shape parameterization and flow condition; this illustration primarily highlights
the relationship between BO, NN, and data collection.
A more detailed explanation of the input and output quantities will be provided at the beginning of \cref{sec:nn}.

This paper is structured as follows. In \cref{sec:nn}, we demonstrate the problem formulation and baseline NN design. In \cref{sec:bo}, we illustrate the process of using Bayesian optimization to tune the NN HP. And in \cref{sec:result}, we demonstrate the accuracy and other metrics of the optimized NN model. Finally, \cref{sec:conclusion} concludes this research.

\section{Baseline Neural Network Design}
\label{sec:nn}
In this section, we aim to build a model that approximate the performance of an airfoil with 14 design variables and two flow conditions. 
The model predicts two quantities, \textit{i.e.}, the lift coefficient $\mathrm{C_L}$ and the drag coefficient $\mathrm{C_D}$, that are used to quantify the aerodynamic performance.
A modal parameterization approach is employed to parameterize the airfoil geometry, where the shape is defined as a linear combination of a finite number of airfoil modes, using a set of ``modes weights'' as the coefficients. In other words, the airfoil geometry is constructed by  the dot product between weights and modes~\cite{grey2018active}.
The two flow conditions are the Mach number and the angle-of-attack which also impact the two coefficients $\mathrm{C_L}$ and $\mathrm{C_D}$. This workflow is depicted in~\cref{fig:ASO-BO}.
The dataset contains \numprint{42039} samples, each with 16 inputs and two output variables.
Separate models are used to predict each output.
For demonstration purposes, only the model for $\mathrm{C_D}$ is presented. 
Still, the same approach applies to the the $\mathrm{C_L}$ model.

\begin{figure}[!ht]
    \centering
    \includegraphics[width=\linewidth]{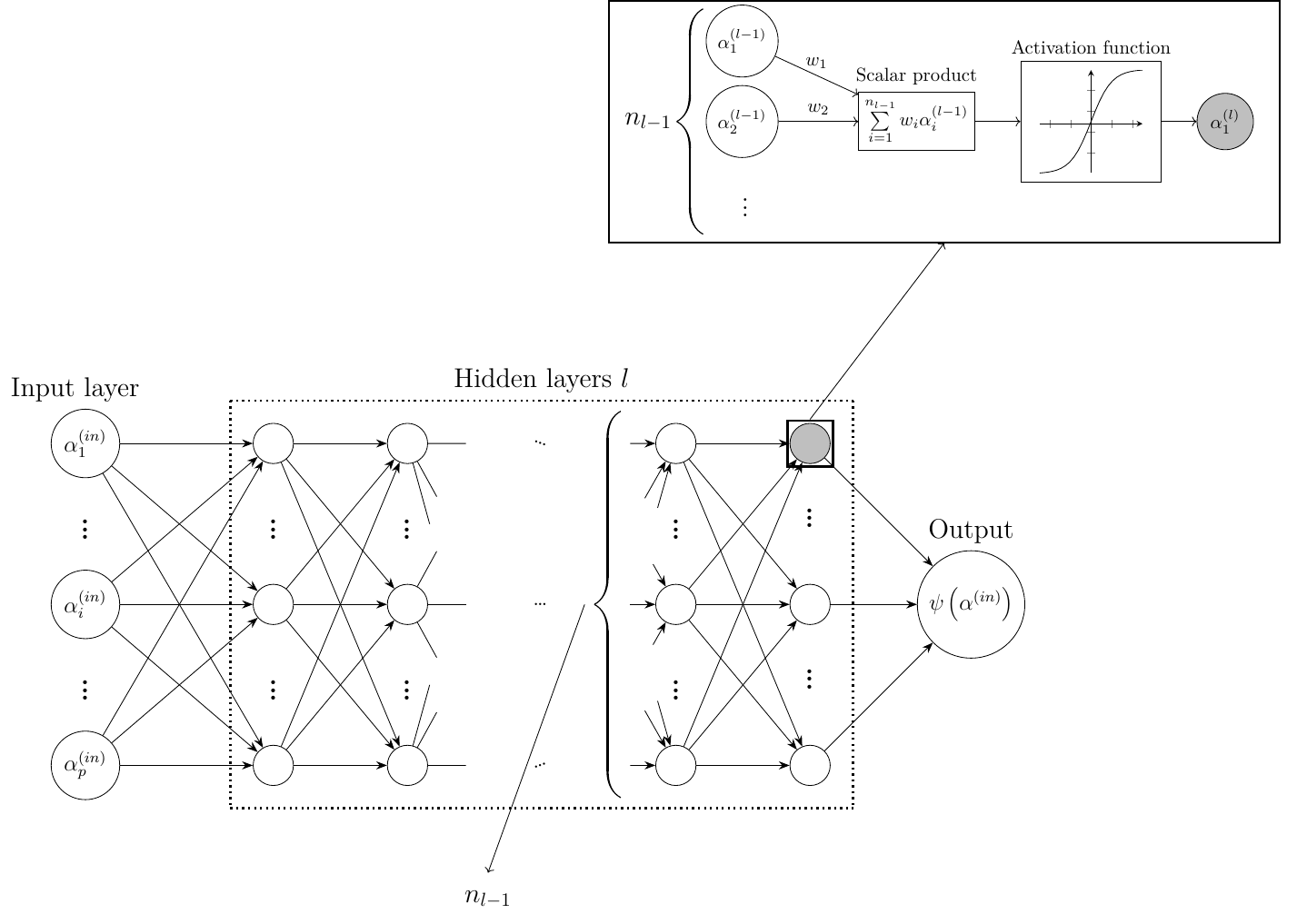}
    \caption{A sample plot showing an MLP architecture. The shown number of nodes is only for illustration (figure adapted from~\cite[Figure 1]{audet2022general}).}
    \label{fig:nn-arch}
\end{figure}

A shallow Multi-Layer Perceptron (MLP) model is chosen as the baseline for several reasons. 
\cref{fig:nn-arch} shows a sample MLP architecture. In this plot, the shown number of nodes is arbitrarily selected for illustration purposes only; in real applications, the model typically uses more nodes.
First, the number of neurons per layer can be flexibly scaled based on the number of samples available and on the dimension of the problem. 
Second, the using fewer layers reduces computational cost, which is important for large-scale MDO problems where a large number of function evaluations are required. 
Furthermore, MLP serves as universal approximators, making them well-suited for complex modeling applications~\cite{hornik1989multilayer}.
Amongst all \numprint{42039} samples, 80\% are used for training, 10\% for validation, and 10\% for testing.
The validation samples are used to evaluate the model performance by Cross Validation (CV).
When the CV Mean Squared Error (MSE) stops decreasing for a few epochs, the training process stops to avoid overfitting.
Finally, the CVMSE on the test set is used as its performance metrics.

To minimize the training loss with backpropagation, the Levenberg-Marquardt (LM) algorithm is employed. This algorithm combines the advantages of gradient descent and the Gauss-Newton method for efficient network parameter optimization~\cite{marquardt1963algorithm}, which makes it particularly effective for training moderate-sized neural networks, achieving second-order convergence speed~\cite{hagan1994training}.
Primarily, we want to minimise total MSE loss
\begin{equation}
    L\left(\beta\right) = \sum\limits_{n = 1}^N \norm{{y_n} - f\left({x_n};\beta\right)},
\end{equation}
where $\beta$ is the model parameter, $x_n$ and $y_n$ denote individual input and loss, and $n=1,2,3,\ldots,N$ is the index of the training sample.
Let individual loss item be expressed as
\begin{equation}
    {r_n}(\beta ) = \norm{{y_n} - f({x_n};\beta)}_2,
\end{equation}
we then have the Hessian matrix
\begin{equation}
    \vec{H}=\pdv{L(\beta)}{\beta_i,\beta_j}=2\sum\limits_{n = 1}^N\qty(\pdv{r_n}{\beta_i}\pdv{r_n}{\beta_j}+r_n\pdv{r_n}{\beta_i,\beta_j})=2\sum\limits_{n = 1}^N\qty(J_n^TJ),
\end{equation}
where
\begin{equation}
    J_n=\pdv{r_n}{\beta_i}
\end{equation}
is the Jacobian matrix of the $n$-th loss.
In LM algorithm, the Hessian matrix is approximated by
\begin{equation}
    \vec{H}=\vec{J}^T\vec{J},
\end{equation}
and the cross derivative term is ignored.
With this simplification, the LM update rule for trainable parameters can be written in vector form as
\begin{equation}
    \vec{\beta }_{k + 1} = \vec{\beta }_k - \vec{H}^{-1}\vec{g} = \vec{\beta }_k - \left[ \vec{J}^T\vec{J} + \mu \vec{I} \right]^{ - 1}\vec{J}^T\vec{r},
    \label{eq:lm-update}
\end{equation}
where
\begin{equation}
    \vec{g}=\pdv{L}{\beta}=\vec{J}^T\vec{r}
\end{equation}
is the gradient.
When $\mu\gg1$, \cref{eq:lm-update} becomes gradient descent with a small step size. 
The damping factor $\mu$ is repetitively decreased after each successful step ($L$ decreases), to recover the Gauss-Newton method.

\section{Bayesian Optimization}
\label{sec:bo}
Traditional optimization methods, such as gradient descent or genetic algorithms, can be computationally prohibitive when the objective function is costly to compute, lacks derivatives, or is highly non-convex~\cite{greenhill2020bayesian}. In this setting, Bayesian Optimization (BO) is a stochastic optimization technique that has gained more attention in recent years, particularly for optimizing functions that are expensive to evaluate and derivative free~\cite{AuHa2017}. Being a black-box optimization algorithm, BO does not require an explicit form of the expensive function $f$. Instead, it builds a surrogate model---where a Gaussian Process (GP) is commonly employed---to approximate the objective function based on a limited number of evaluations ($X_t, f(X_t)$) called a Design of Experiments (DoE)~\cite{williams1995gaussian}.
In this work, we target to learn an inexpensive surrogate model $ \hat{f}$ from a mixed-categorical black-box function given by
\begin{equation}
f :  \Omega \times S \times \mathbb{F}^l \to \mathbb{R}.
  \label{eq:opt_prob}
\end{equation}
This function $f$ is typically an expensive-to-evaluate simulation with no exploitable derivative information.
 $\Omega \subset \mathbb{R}^n$ represents the bounded continuous design set for the $n$ continuous variables.  $S \subset \mathbb{Z}^m$ represents the bounded integer set where $L_1, \ldots, L_m$ are the numbers of levels of the $m$ quantitative integer variables on which we can define an order relation and $ \mathbb{F}^l = \{1, \ldots, L_1\} \times \{1, \ldots, L_2\} \times  \ldots \times \{1, \ldots, L_l\}$ is the design space for the $l$ categorical qualitative variables with their respective  $L_1, \ldots, L_l$ levels.
Namely,
we will consider that our unknown black-box function $f$ follows a Gaussian process of mean $\mu^{f}$ and of standard deviation $\sigma^f$, $\textit{i.e.}$,
\begin{equation}
f \sim \hat{f}=
\mathcal{GP} \left(\mu^{f}, \left[\sigma^f\right]^2\right). \label{eq:GP:f}\end{equation}
The BO process iteratively builds the surrogate model to improve its accuracy as more data become available. At every iteration, an acquisition function is used to balance exploration (searching unexplored regions) and exploitation (refining known good areas), determining the best next point $\mathbf{x}_{\text{next}}$ to evaluate. The acquisition function is crucial in BO as it determines the efficiency of the search process. Common acquisition functions include Expected Improvement (EI), Probability of Improvement (PI), Upper Confidence Bound (UCB) or the Watson and Barnes $2^{\text{nd}}$ criterion smoothed (WB2s)~\cite{Jones2001JOGO, shahriari2015taking,bartoli2019adaptive}.
As mentioned above, GP is the keystone for BO due to its flexibility and ability to quantify uncertainty. More precisely, a GP is fully specified by its mean function $\hat{\mu}^f$ and correlation kernel $k$, which together define the GP and allows to predict, for any input $x$, a mean prediction $\mu^f(x)$ and a standard deviation $\sigma^f(x)$~\cite{williams1995gaussian}. 
The chosen kernel $k$ significantly impacts the performance of the GP in modeling the underlying function. The most used kernels are the exponential and Mat\'ern kernels~\cite{Lee2011}.
One of the key advantages of BO is its ability to efficiently handle expensive-to-evaluate objective functions with a limited number of evaluations. This makes BO particularly suitable for applications in hyperparameter tuning of machine learning models, automated design of experiments or engineering systems optimization~\cite{frazierTutorialBayesianOptimization2018, snoek2015scalable}. 
BO methods have also been adapted for multi-objective optimization, optimization under constraints, handling mixed continuous and categorical variables, and even high-dimensional search spaces by incorporating dimensionality reduction techniques like Partial Least Squares (PLS) regression~\cite{grapin_constrained_2022, Mixed_Paul_PLS, Mixed_Paul,SEGO-UTB}. 
The approach for modeling categorical variables is based on the work by Saves~\textit{et al.}~\cite{Mixed_Paul}, which uses dedicated kernels to model categorical variables, for example the Gower Distance (GD), the Continuous Relaxation (CR), the Compound Symmetry (CS), Homoscedastic Hypersphere (HH) or Exponential Homoscedastic Hypersphere (EHH) kernel. In particular, we use the implementation in the Surrogate Modeling Toolbox (SMT)\footnote{\label{note:smt}\href{https://smt.readthedocs.io/}{https://smt.readthedocs.io/}}~\cite{SMT2}, where the CR kernel is used by default.  
One of the main limitations of BO is that GP models do not handle high-dimensional design spaces well. Due to the number of hyper-parameters to tune, training and sampling time escalates rapidly with the numbers of input dimensions and training points~\cite{garnett2023bayesian}. However, a feature space can be defined that reduces the dimension of the original input space~\cite{calandra2016bayesian}, supported by the observation that the optimization problem has a lower intrinsic dimensionality in some cases.
Our BO algorithm uses Kriging with Partial Least Squares (KPLS) to construct such feature spaces as described in~\cite{Bouhlel18}. Recently, KPLS has been extended to work with discrete and hierarchical variables too~\cite{Mixed_Paul_PLS,charayron2023towards,SMT2} and these methods are implemented in SMT~\cite{SMT,SMT2}. The latter work incorporates \texttt{Categorical}, \texttt{Integer}, \texttt{Ordinal} and \texttt{Continuous} types of variables that cover \texttt{Meta}, \texttt{Meta-Decreed}, \texttt{Decreed} and \texttt{Neutral} roles of variables~\cite{bussemaker2024system,halle2024graph}.
By default, KPLS is applied with $n_{\mathrm{kpls}} = 10$ if the design space contains more than ten design variables.
For an arbitrary $w= (x,z,c) \in \mathbb{R}^n \times  \mathbb{Z}^m \times \mathbb{F}^l$, not necessary in the DoE, the GP model prediction at $w$ writes as $\hat f(w) = \mu ({w})+\epsilon(w) \in \mathbb{R} $, with $\epsilon$ being the error between $f$ and the model approximation $\mu$~\cite{GP14}. The considered error terms are random variables of variance $\sigma^2$.  Using the DoE, the expression of  $\mu^{f}$ and the estimation of its variance $[\sigma^f]^2$ are given as follows:

\begin{equation} \label{eq:mean:GP}
\mu^f\left(w\right)= \hat{\mu}^f+r\left(w\right)^\top  \left[R\left(\Theta\right)\right]^{-1}\left(\textbf{y}^f-\mathds{1} \hat{\mu}^f\right), 
\end{equation}
and
\begin{equation}
\label{eq:std:GP}
\left[\sigma^f\left(w\right)\right]^2=\left[\hat{\sigma}^f\right]^2\left[1-r\left(w\right)^\top  \left[R\left(\Theta\right)\right]^{-1}r\left(w\right)+ \frac{\left(1-\mathds{1}^\top  [R(\Theta)]^{-1}r(w) \right)^2}{\mathds{1}^\top  \left[R\left(\Theta\right)\right]^{-1}\mathds{1}}\right], \end{equation}
where $\hat{\mu}^f$ and $\hat{\sigma}^f$ are the maximum likelihood estimator (MLE)~\cite{MLE} of $\mu$ and $\sigma$, respectively. $\mathds{1}$ denotes the vector of $n_t$ ones. $R$ is the $ n_t \times n_t $ correlation matrix between the input points and $r(w)$ is the correlation vector between the input points and a given $w$.
The correlation matrix $R$ is defined, for a given couple  $\left(r,s\right) \in \left(\left\{1,\ldots,n_t\right\}\right)^2$, by 
\begin{equation}
    \label{eq:R}
    \left[R\left(\Theta\right)\right]_{r,s}=k\left(w^r,w^s,\Theta\right) \in \mathbb{R},
\end{equation}
and the vector $r\left(w\right)\in \mathbb{R}^{n_t}$ is defined as $r\left(w\right) =\left[k\left(w,w^1\right), \ldots , k\left(w,w^{n_t}\right)\right]^{\top}$,
where $k$ is a given correlation kernel that relies on a set of hyper-parameters $\Theta$. 
The mixed-categorical  correlation kernel is given as the product of three kernels:
\begin{equation}
k\left(w^r,w^s,\Theta\right) =  k^\mathrm{cont}\left(x^r,x^s,\theta^\mathrm{cont}\right) \cdot k^\mathrm{int}\left(z^r,z^s,\theta^\mathrm{int}\right)
\cdot k^\mathrm{cat}\left(c^r,c^s,\theta^\mathrm{cat}\right),
\label{eq:decomp_mix}
\end{equation}
where $k^\mathrm{cont}$  and $\theta^\mathrm{cont}$ are the continuous kernel and its associated hyper-parameters, $k^\mathrm{int}$  and $\theta^\mathrm{int}$ are the integer kernel and its hyper-parameters, and lastly $k^\mathrm{cat}$  and $\theta^\mathrm{cat}$ are the ones related with the categorical inputs. In this case, one has $\Theta=\{ \theta^\mathrm{cont},\theta^\mathrm{int},\theta^\mathrm{cat}\}$. 
Henceforth, the general correlation matrix $R$ will rely only on the set of the hyper-parameters $\Theta$:
\begin{equation}
    \label{eq:corel:mat}
    \left[R\left(\Theta\right)\right]_{r,s} = \left[R^\mathrm{cont}\left(\theta^\mathrm{cont}\right)\right]_{r,s} \cdot
    \left[R^\mathrm{int}\left(\theta^\mathrm{int}\right)\right]_{r,s}
    \cdot
    \left[R^\mathrm{cat}\left(\theta^\mathrm{cat}\right)\right]_{r,s},
\end{equation}
where $\left[R^\mathrm{cont}\left(\theta^\mathrm{cont}\right)\right]_{r,s} =k^\mathrm{cont}\left(x^r,x^s,\theta^\mathrm{cont}\right)$, $\left[R^\mathrm{int}\left(\theta^\mathrm{int}\right)\right]_{r,s} =k^\mathrm{int}\left(z^r,z^s,\theta^\mathrm{int}\right)$ and  $\left[R^\mathrm{cat}\left(\theta^\mathrm{cat}\right)\right]_{r,s}=k^\mathrm{cat}\left(c^r,c^s,\theta^\mathrm{cat}\right)$. 
The set of hyper-parameters $\Theta$ could be estimated using the DoE data set $({W},\textbf{y}^f)$ through the MLE approach on the following way
\begin{equation}
\Theta^*= \arg\max_{\Theta} \mathcal{L}\left(\Theta\right):=\left( - \frac{1}{2} {\textbf{y}^f}^\top \left[R\left(\Theta\right)\right]^{-1} {\textbf{y}^f}   - \frac{1}{2} \log 	\abs{  \left[R\left(\Theta\right)\right]} - \frac{n_t}{2} \log 2 \pi    \right),
\label{eq:likelihood}
\end{equation}
where $R\left(\Theta\right)$ is computed using Eq.~\ref{eq:corel:mat}. 

To optimize the NN design, there are five HPs to be tuned by Bayesian optimization, as listed in \cref{tab:hyp}.
$N$ can take any positive integer values.
The rest of the variables need to have ``discrete'' values; $N_1$, $N_2$, and $N_3$ are ordinal variables because their values follow a strict order with a fixed interval between them, whereas $F$ is a categorical variable.
$N$ is a meta-variable which controls the behavior of the decreed variable $N_3$, \textit{i.e.}, $N_3$ is only activated when $N=3$.

\begin{table}[ht!]
\centering
\begin{tabular}{lrrrrr}
\toprule
Hyperparameter                     & Lower Bound    & Upper Bound    & Interval   & Type        & Role    \\
\midrule
Number of layers, $N$               & 2              & 3              & 1         & Integer     & Meta    \\
Number of neurons in layer 1, $N_1$ & 10             & 80             & 5         & Ordinal     & Neutral \\
Number of neurons in layer 2, $N_2$ & 10             & 80             & 5         & Ordinal     & Neutral \\
Number of neurons in layer 3, $N_3$ & 10             & 80             & 5         & Ordinal     & Decreed \\
Activation Function, $F$            & \multicolumn{3}{c}{\{\texttt{ReLU}, \texttt{tanh}, \texttt{sigmoid}\}}                  & Categorical              & Neutral                 \\
\bottomrule
\end{tabular}
\caption{Tunable HPs in the NN design and their value ranges.}
\label{tab:hyp}
\end{table}

\section{Results}
\label{sec:result}
In this section, we demonstrate the benefits of the developed framework to tune NN architecture, by comparing the Bayesian-optimized NN HPs against the baseline design and other publicly available NNs.

\subsection{Performance Metrics}

To evaluate the accuracy of the predictive models, we employ several commonly used error metrics: Mean Square Error (MSE), Root Mean Square Error (RMSE), and Mean Absolute Percentage Error (MAPE). In addition, to describe the cost-effectiveness of the model, we propose Parameter Efficiency (PE). These metrics are defined mathematically as follows:

\paragraph{MSE} \hspace{-9pt} measures the average of the squared differences between the predicted values $\hat{y}_i$ and the true values $y_i$. MSE emphasizes larger errors due to the squaring operation and is useful for highlighting significant deviations in predictions. It is given by:
\begin{equation}
\mathrm{MSE} = \frac{1}{n} \sum_{i=1}^{n} \left( y_i - \hat{y}_i \right)^2,
\end{equation}
where $n$ is the total number of data points, $y_i$ represents the actual value, and $\hat{y}_i$ represents the predicted value.

\paragraph{RMSE} \hspace{-9pt} is the square root of the MSE and provides an interpretable error metric in the same units as the original data:
\begin{equation}
\mathrm{RMSE} = \sqrt{\mathrm{MSE}} = \sqrt{\frac{1}{n} \sum_{i=1}^{n} \left( y_i - \hat{y}_i \right)^2}.
\end{equation}

\paragraph{MAPE} \hspace{-9pt} measures the average absolute percentage difference between predicted and true values, making it a scale-independent metric:
\begin{equation}
\mathrm{MAPE} = \frac{1}{n} \sum_{i=1}^{n} \left| \frac{y_i - \hat{y}_i}{y_i} \right| \times 100,
\end{equation}
where the result is expressed as a percentage. MAPE is particularly useful when comparing model performance across datasets with varying scales.

\paragraph{PE} \hspace{-9pt} evaluates the accuracy achieved per parameter. Granted, a larger model tends to perform better, however, is more computationally expensive. We define
\begin{equation}
    \mathrm{PE}=\frac{\text{Accuracy}}{\text{Number of parameters}}.
\end{equation}
As accuracy is typically only defined for categorization tasks rather than regression tasks which is our case, we define a reasonable PE for this task by
\begin{equation}
    \text{PE}=\frac{\max\qty(1-\zeta~\mathrm{MAPE},0)}{\text{Number of parameters}},
\end{equation}
where $\zeta$ is a tolerance factor, which scales MAPE based on the particular problem.
In this problem, we set $\zeta=100$, which means when MAPE goes beyond 1\%, the model losses all its usability.

\subsection{Optimization of the Baseline NN}
The baseline network has two layers with 20 neurons per layer and \texttt{tanh} activation function. 
As the objective function exhibits a relatively small magnitude, it is normalized using $z$-score normalization to achieve a mean of unity. Recall that amongst all \numprint{42039} samples, 80\% are used for training, 10\% for validation, and 10\% for testing.
The performance of the baseline model is measured by a CV MSE of \numprint{3.06e-7} and a CV MAPE of \numprint{0.1433}\%. 
Our experimental results indicate that this level of accuracy is attributable to Jacobian training in LM and cannot be attained using first-order optimization methods, such as Stochastic Gradient Descent (SGD).
This performance is comparable with the results presented by Li~\textit{et al.}~\cite{Li2019}, but the model is significantly lighter\footnote{Model parameters occupy a storage of 15 kilobytes and the training time is 54 seconds on a personal laptop.}.

\begin{figure}[ht!]
    \centering
    \includegraphics[width=0.7\linewidth]{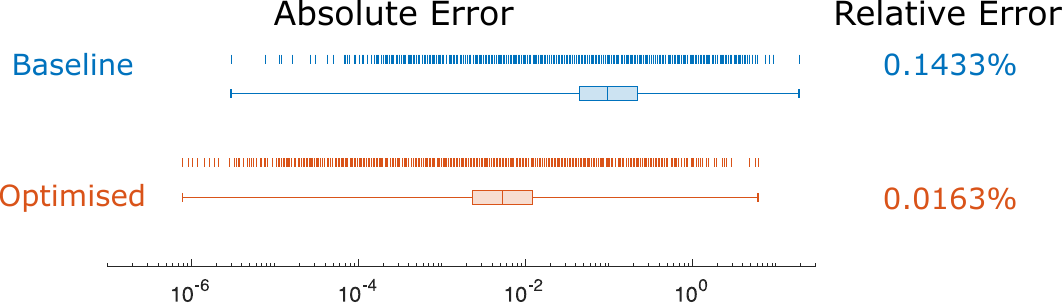}
    \caption{Error distribution of the baseline and optimized model.}
    \label{fig:error-dist}
\end{figure}

The Bayesian-optimized NN has three layers with 50, 50, 60 neurons each and \texttt{tanh} activation function.
The optimized model exhibits a CV MSE of \numprint{8.14e-8} and a CV MAPE of only \numprint{0.0163}\%, which is about one order of magnitude smaller than the baseline. 
\cref{fig:error-dist} plots the error distributions of the baseline and optimized models. On the left, the figure shows absolute errors measured in drag count (equivalent to \numprint{1e-4} $\mathrm{C_D}$), while the MAPE values of both models are shown on the right.

\begin{figure}[ht!]
    \centering
    \includegraphics[width=0.6\linewidth]{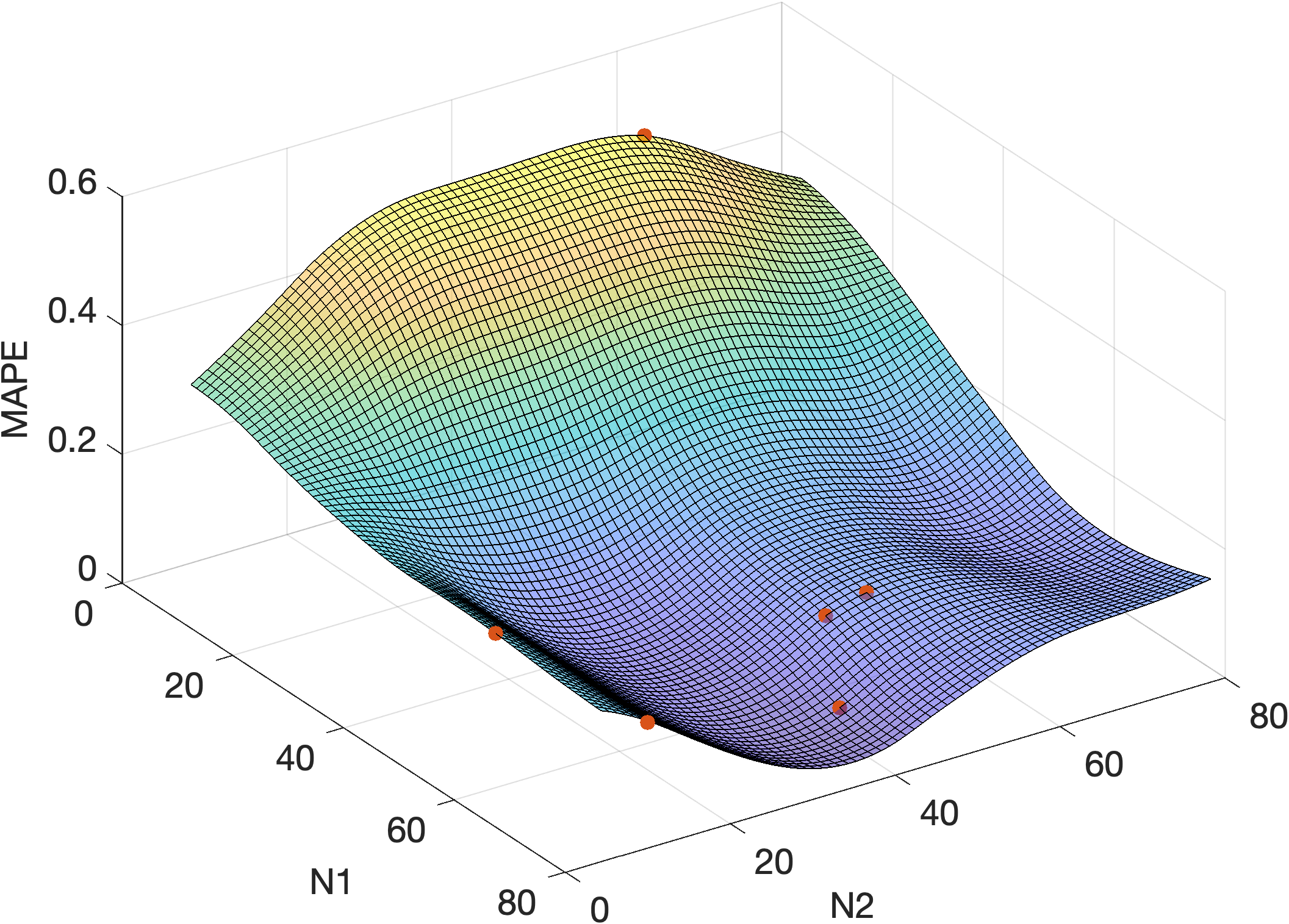}
    \caption{GP's prediction on MAPE given $N_1$ and $N_2$ with all other variables fixed, where solid red dots are collected samples during the BO process.}
    \label{fig:slice}
\end{figure}

In order to evaluate and understand the underlying GP model during the BO process, we slice the 5-D input data (with five features) into 2-D spaces and plot the predicted MAPE values.
In particular, we fix $N=2$ (and hence, $N_3$ is empty because there is no third layer),
and $F$ is set to \texttt{sigmoid}.
\cref{fig:slice} depicts the MAPE in response to varying $N_1$ and $N_2$. 
Solid red dots are collected samples during the BO process and the surface is the GP prediction of MAPE, {where the color corresponds to the MAPE values shown on the $z$-axis.}
We see that the GP model interpolates the data with a reasonable curvature, which indicates that GP is neither overfitting nor underfitting.

\subsection{Comparison with Other NN Models}
\texttt{NeuralFoil}~\cite{neuralfoil} is a well-known open-source implementation of MLP NN for aerodynamic performance prediction, which is available as a Python library.
\texttt{NeuralFoil} contains eight model sizes, with the smallest model having a single hidden layer of 48 neurons, while the largest model has five hidden layers of 512 neurons each.
We select and train five models to compare with our Bayesian-optimized NN model.
Both MAPE and PE metrics are selected for comparison to measure its absolute performance and cost-effectiveness.
\begin{table}[]
\centering
\caption{Performance comparison between of Neural Foil and Bayesian-optimized NN.}
\label{table:comparison}
\begin{tabular}{lllllll}
\toprule
Model Size       & \texttt{xxsmall} & \texttt{xsmall} & \texttt{small} & \texttt{BO}   & \texttt{medium} & \texttt{large} \\
\midrule
Number of Layers & 1 & 2 & 2 & 3 & 3 & 3\\
Number of Parameters    & \numprint{865} & \numprint{3217} & \numprint{5313} & \numprint{6521} & \numprint{9473} & \numprint{35329}\\
MAPE             &  \numprint{0.141} & \numprint{0.088} & \numprint{0.124} & \numprint{0.016} & \numprint{0.410} & \numprint{1.053}\\
PE $(\times10^3)$ & \numprint{0.993} & \numprint{0.284} & \numprint{0.165} & \numprint{0.090} & \numprint{0.000} & \numprint{0.000}
\\
\bottomrule
\end{tabular}
\end{table}

The smallest models, consisting of one or two layers, are observed to have lower accuracy. 
This can be attributed to their insufficient capacity to capture the complexity inherent in the underlying data. 
An intriguing finding is that the \texttt{xsmall} model (referring to Table~\ref{table:comparison}) outperforms the \texttt{small} model in our tests. 
Among the three 3-layer models, the BO model achieves the best performance, despite being the smallest of the three, showing an accuracy improvement of nearly an order of magnitude over the smallest model. 
However, further increasing the number of parameters results in degraded performance, because optimizing a larger number of parameters poses significant challenges, ultimately hindering the model's ability to generalize effectively\footnote{Theoretically, if the optimizer can find the true global optimum for all trainable parameters, the performance of a larger MLP is always at least equal to the smaller one. This is a result of the Universal Approximation Theorem~\cite{cybenko1989approximation}.}.

We observe a monotonic decline in PE as the number of model parameters increases. 
Notably, the smallest model exhibits a significantly higher PE compared to the larger models. 
This behavior is expected, as increasing the model size beyond a certain point yields only marginal improvements in performance. 
When the model becomes excessively large, its accuracy begins to deteriorate, resulting in a substantial reduction in PE.

\subsection{Additional Benchmark Problem: Aircraft Self Noise}
An additional benchmark is selected to further demonstrate the effectiveness of this BO formulation.
The task is to predict the noise generated by an aircraft from its operating conditions.
The data are publicly available on \footnote{Accessed on 5 Feb 2025: \url{https://www.kaggle.com/datasets/fedesoriano/airfoil-selfnoise-dataset}.}{\href{https://www.kaggle.com/datasets/fedesoriano/airfoil-selfnoise-dataset}{Kaggle}} website, where explanation of each input and output variable is detailed.
We follow exactly the same workflow as discussed in \cref{sec:nn,sec:bo}.
As the dataset is smaller compared with our aerodynamic performance data, each layer is limited with fewer neurons.
\cref{tab:hyp-kaggle} tabulates the configuration of tunable HPs in this problem.

\begin{table}[ht!]
\centering
\begin{tabular}{lrrrrr}
\toprule
Hyperparameter                     & Lower Bound    & Upper Bound    & Inteval   & Type        & Role    \\
\midrule
Number of layers $N$               & 2              & 3              & 1         & Ordinal     & Meta    \\
Number of neurons in layer 1 $N_1$ & 5             & 40             & N/A         & Integer     & Neutral \\
Number of neurons in layer 2 $N_2$ & 5             & 40             & N/A         & Integer     & Neutral \\
Number of neurons in layer 3 $N_3$ & 5             & 40             & N/A         & Integer     & Decreed \\
Activation Function $F$            & \multicolumn{3}{c}{\{\texttt{ReLU}, \texttt{tanh}, \texttt{sigmoid}\}}                  & Categorical              & Neutral                 \\
\bottomrule
\end{tabular}
\caption{Tunable HP in the NN design for the Kaggle benchmark problem.}
\label{tab:hyp-kaggle}
\end{table}

The optimized NN has three layers with 40, 24, 37 neurons each and its activation function is \texttt{sigmoid}.
The CV MAPE is merely 0.82\% and CV MSE is \SI{2.3476}{\dB}.
We compare our result with the most popular answer submitted to Kaggle by K. Majdi\footnote{Accessed on 5 Feb 2025: \url{https://www.kaggle.com/code/majdikarim/ann-linear-regression-polynomial-regression}.}.
Three models are used in his submission as shown in~\cref{tab:compare-kaggle}.
It can be clearly seen that the NN obtained by BO has a much better accuracy with substantially smaller footprint.

\begin{table}[ht]
\centering
\begin{tabular}{r|rrr|r}
\toprule
 & \multicolumn{3}{c|}{Kaggle} & \textbf{BO} \\
 & NN & Linear Regression & Polynomial Regression & NN \\
\midrule
Specs & \begin{tabular}[c]{@{}r@{}}$N=2$\\ $(300,300)$\end{tabular} &                   &                       & \begin{tabular}[c]{@{}r@{}}$N=3$\\ $(40,24,47)$\end{tabular} \\
MAPE  & 3.40\% & 2.43\% & 3.36\% & \textbf{0.82\%}\\
\bottomrule
\end{tabular}
\caption{Comparison between Kaggle submission and BO result.}
\label{tab:compare-kaggle}
\end{table}

\section{Conclusion}
\label{sec:conclusion}
In this study, we have demonstrated the ability of BO in tuning the HPs of NN for aerodynamic performance prediction and aircraft self-noise estimation. 
In addition to conventional BO formulation, specialized kernels are used to fit each variable role (meta or decreed) and characteristics (ordinal, integer, or categorical), where the HP space is efficiently explored.
Trainable parameters within NN are optimized by LM backpropagation for improved convergence rate, since the problem size is suitable for a 2\textsuperscript{nd}-order Hessian-required optimization method.
Quantitatively, the optimized NN for aerodynamic performance prediction achieved a cross-validation MAPE of just 0.0163\%, offering an order of magnitude enhancement over the baseline model's MAPE of 0.1433\%. 
Furthermore, the optimized model outperformed other publicly available NN models in terms of both accuracy and cost-effectiveness.
In the aircraft self-noise estimation benchmark, our Bayesian-optimized neural network achieved a MAPE of 0.82\%, significantly outperforming the most popular Kaggle submission, which reported a MAPE of 3.40\%. 
These benchmark problems highlight the robustness and adaptability of our method across different datasets and problem domains.
To offer more flexibility on the NN design, it is possible to fuse different activation functions to form a generic one by weighted average, where additional HPs are introduced.
This will be one of the directions of our ongoing research.

\section*{Acknowledgement}

We would like to acknowledge the support from the University Grant Council of the Hong Kong Special Administration Region for providing financial support to the first author J. M. Shihua through the Hong Kong PhD Fellowship Scheme, grant number PF20-42345.

This work is part of the activities of ONERA - ISAE - ENAC joint research group.
The research presented in this paper has been performed within the framework of the COLOSSUS project (Collaborative System of Systems Exploration of Aviation Products, Services and Business Models) and has received funding from the European Union Horizon Programme under grant agreement n${^\circ}$101097120. The authors acknowledge the research project MIMICO funded in France by the Agence Nationale de la Recherche (ANR, French National Research Agency), grant number ANR-24-CE23-0380.

\newpage
\clearpage
\setlength{\bibsep}{2pt}
\bibliography{main.bib}

\begin{thebibliography}{43}
\newcommand{\enquote}[1]{``#1''}
\providecommand{\natexlab}[1]{#1}
\providecommand{\url}[1]{\texttt{#1}}
\providecommand{\urlprefix}{URL }
\expandafter\ifx\csname urlstyle\endcsname\relax
  \providecommand{\doi}[1]{\discretionary{}{}{}https://doi.org/#1}\else
  \providecommand{\doi}[1]{\discretionary{}{}{}\urlstyle{rm}\url{https://doi.org/#1}}\fi

\bibitem[{Martins and Lambe(2013)}]{martins2013multidisciplinary}
Martins, J.~R., and Lambe, A.~B., \enquote{Multidisciplinary design optimization: a survey of architectures,} \emph{AIAA journal}, Vol.~51, No.~9, 2013, pp. 2049--2075.

\bibitem[{Bouhlel et~al.(2019)Bouhlel, Hwang, Bartoli, Lafage, Morlier, and Martins}]{SMT}
Bouhlel, M.~A., Hwang, J.~T., Bartoli, N., Lafage, R., Morlier, J., and Martins, J.~R., \enquote{A Python surrogate modeling framework with derivatives,} \emph{Advances in Engineering Software}, Vol. 135, 2019.
\newblock \doi{10.1016/j.advengsoft.2019.03.005}.

\bibitem[{Cressie(1990)}]{OK}
Cressie, N., \enquote{The origins of kriging,} \emph{Mathematical Geology}, Vol.~22, No.~3, 1990, p. 239–252.
\newblock \doi{10.1007/bf00889887}.

\bibitem[{Bouhlel et~al.(2016)Bouhlel, Bartoli, Otsmane, and Morlier}]{KPLS}
Bouhlel, M.~A., Bartoli, N., Otsmane, A., and Morlier, J., \enquote{Improving kriging surrogates of high-dimensional design models by Partial Least Squares dimension reduction,} \emph{Structural and Multidisciplinary Optimization}, Vol.~53, No.~5, 2016, p. 935–952.
\newblock \doi{10.1007/s00158-015-1395-9}.

\bibitem[{Bouhlel and Martins(2018)}]{GEKPLS}
Bouhlel, M.~A., and Martins, J. R. R.~A., \enquote{Gradient-enhanced kriging for high-dimensional problems,} \emph{Engineering with Computers}, Vol.~35, No.~1, 2018, p. 157–173.
\newblock \doi{10.1007/s00366-018-0590-x}.

\bibitem[{Bauer et~al.(2016)Bauer, van~der Wilk, and Rasmussen}]{SGP}
Bauer, M., van~der Wilk, M., and Rasmussen, C.~E., \enquote{Understanding probabilistic sparse Gaussian process approximations,} \emph{Proceedings of the 30th International Conference on Neural Information Processing Systems}, Curran Associates Inc., Red Hook, NY, USA, 2016, p. 1533–1541.

\bibitem[{Hwang and Martins(2018)}]{RMTS}
Hwang, J.~T., and Martins, J.~R., \enquote{A fast-prediction surrogate model for large datasets,} \emph{Aerospace Science and Technology}, Vol.~75, 2018, pp. 74--87.
\newblock \doi{10.1016/j.ast.2017.12.030}.

\bibitem[{Bouhlel et~al.(2020)Bouhlel, He, and Martins}]{GENN}
Bouhlel, M.~A., He, S., and Martins, J. R. R.~A., \enquote{Scalable gradient–enhanced artificial neural networks for airfoil shape design in the subsonic and transonic regimes,} \emph{Structural and Multidisciplinary Optimization}, Vol.~61, No.~4, 2020, p. 1363–1376.
\newblock \doi{10.1007/s00158-020-02488-5}.

\bibitem[{Sharpe and Hansman(2025)}]{neuralfoil}
Sharpe, P.~R., and Hansman, R.~J., \enquote{{NeuralFoil}: An airfoil aerodynamics analysis tool using physics-informed machine learning,} \emph{arXiv preprint arXiv:2503.16323}, 2025.

\bibitem[{Bengio(2012)}]{bengio2012practical}
Bengio, Y., \enquote{Practical recommendations for gradient-based training of deep architectures,} \emph{Neural networks: Tricks of the trade: Second edition}, Springer, 2012, pp. 437--478.

\bibitem[{Bartz-Beielstein(2023)}]{bartz2023hyperparameter}
Bartz-Beielstein, T., \enquote{Hyperparameter Tuning Cookbook: A guide for scikit-learn, PyTorch, river, and spotPython,} \emph{arXiv preprint arXiv:2307.10262}, 2023.

\bibitem[{Akiba et~al.(2019)Akiba, Sano, Yanase, Ohta, and Koyama}]{optuna}
Akiba, T., Sano, S., Yanase, T., Ohta, T., and Koyama, M., \enquote{Optuna: A Next-generation Hyperparameter Optimization Framework,} \emph{Proceedings of the 25th {ACM} {SIGKDD} International Conference on Knowledge Discovery and Data Mining}, 2019, pp. 2623--2631.

\bibitem[{Bergstra et~al.(2013)Bergstra, Yamins, and Cox}]{hyperopt}
Bergstra, J., Yamins, D., and Cox, D., \enquote{Making a science of model search: Hyperparameter optimization in hundreds of dimensions for vision architectures,} \emph{International conference on machine learning}, PMLR, 2013, pp. 115--123.

\bibitem[{Snoek et~al.(2012)Snoek, Larochelle, and Adams}]{snoek2012practical}
Snoek, J., Larochelle, H., and Adams, R.~P., \enquote{Practical bayesian optimization of machine learning algorithms,} \emph{Advances in neural information processing systems}, Vol.~25, 2012.

\bibitem[{Saves et~al.(2024{\natexlab{a}})Saves, Lafage, Bartoli, Diouane, Bussemaker, Lefebvre, Hwang, Morlier, and Martins}]{SMT2}
Saves, P., Lafage, R., Bartoli, N., Diouane, Y., Bussemaker, J., Lefebvre, T., Hwang, J.~T., Morlier, J., and Martins, J. R. R.~A., \enquote{{SMT 2.0: A} Surrogate Modeling Toolbox with a focus on Hierarchical and Mixed Variables Gaussian Processes,} \emph{Advances in Engineering Sofware}, Vol. 188, 2024{\natexlab{a}}, p. 103571.
\newblock \doi{https://doi.org/10.1016/j.advengsoft.2023.103571}.

\bibitem[{Grey and Constantine(2018)}]{grey2018active}
Grey, Z.~J., and Constantine, P.~G., \enquote{Active subspaces of airfoil shape parameterizations,} \emph{AIAA Journal}, Vol.~56, No.~5, 2018, pp. 2003--2017.

\bibitem[{Audet et~al.(2023)Audet, Hall{\'e}-Hannan, and {Le Digabel}}]{audet2022general}
Audet, C., Hall{\'e}-Hannan, E., and {Le Digabel}, S., \enquote{A general mathematical framework for constrained mixed-variable blackbox optimization problems with meta and categorical variables,} \emph{Operations Research Forum}, Vol.~4, 2023, pp. 1--37.

\bibitem[{Hornik et~al.(1989)Hornik, Stinchcombe, and White}]{hornik1989multilayer}
Hornik, K., Stinchcombe, M., and White, H., \enquote{Multilayer feedforward networks are universal approximators,} \emph{Neural networks}, Vol.~2, No.~5, 1989, pp. 359--366.

\bibitem[{Marquardt(1963)}]{marquardt1963algorithm}
Marquardt, D.~W., \enquote{An algorithm for least-squares estimation of nonlinear parameters,} \emph{Journal of the society for Industrial and Applied Mathematics}, Vol.~11, No.~2, 1963, pp. 431--441.

\bibitem[{Hagan and Menhaj(1994)}]{hagan1994training}
Hagan, M.~T., and Menhaj, M.~B., \enquote{Training feedforward networks with the Marquardt algorithm,} \emph{IEEE transactions on Neural Networks}, Vol.~5, No.~6, 1994, pp. 989--993.

\bibitem[{Greenhill et~al.(2020)Greenhill, Rana, Gupta, Vellanki, and Venkatesh}]{greenhill2020bayesian}
Greenhill, S., Rana, S., Gupta, S., Vellanki, P., and Venkatesh, S., \enquote{Bayesian optimization for adaptive experimental design: A review,} \emph{IEEE access}, Vol.~8, 2020, pp. 13937--13948.

\bibitem[{Audet and Hare(2017)}]{AuHa2017}
Audet, C., and Hare, W., \emph{Derivative-Free and Blackbox Optimization}, Springer Series in Operations Research and Financial Engineering, 2017.

\bibitem[{Williams and Rasmussen(1995)}]{williams1995gaussian}
Williams, C., and Rasmussen, C., \enquote{Gaussian processes for regression,} \emph{Advances in neural information processing systems}, Vol.~8, 1995.

\bibitem[{Jones(2001)}]{Jones2001JOGO}
Jones, D.~R., \enquote{A Taxonomy of Global Optimization Methods Based on Response Surfaces,} \emph{Journal of Global Optimization}, Vol.~21, 2001, pp. 345--383.

\bibitem[{Shahriari et~al.(2015)Shahriari, Swersky, Wang, Adams, and De~Freitas}]{shahriari2015taking}
Shahriari, B., Swersky, K., Wang, Z., Adams, R.~P., and De~Freitas, N., \enquote{Taking the human out of the loop: A review of Bayesian optimization,} \emph{Proceedings of the IEEE}, Vol. 104, No.~1, 2015, pp. 148--175.

\bibitem[{Bartoli et~al.(2019)Bartoli, Lefebvre, Dubreuil, Olivanti, Priem, Bons, Martins, and Morlier}]{bartoli2019adaptive}
Bartoli, N., Lefebvre, T., Dubreuil, S., Olivanti, R., Priem, R., Bons, N., Martins, J.~R., and Morlier, J., \enquote{Adaptive modeling strategy for constrained global optimization with application to aerodynamic wing design,} \emph{Aerospace Science and technology}, Vol.~90, 2019, pp. 85--102.

\bibitem[{Lee(2011)}]{Lee2011}
Lee, H., \emph{{Gaussian} Processes}, Springer Berlin Heidelberg, 2011, Chap.~5, pp. 575--577.

\bibitem[{Frazier(2018)}]{frazierTutorialBayesianOptimization2018}
Frazier, P.~I., \enquote{A {{Tutorial}} on {{{Bayesian} Optimization}},} \emph{ArXiv preprint}, 2018.

\bibitem[{Snoek et~al.(2015)Snoek, Rippel, Swersky, Kiros, Satish, Sundaram, Patwary, Prabhat, and Adams}]{snoek2015scalable}
Snoek, J., Rippel, O., Swersky, K., Kiros, R., Satish, N., Sundaram, N., Patwary, M., Prabhat, M., and Adams, R., \enquote{Scalable {Bayesian} optimization using deep neural networks,} \emph{International conference on machine learning}, 2015, pp. 2171--2180.

\bibitem[{Grapin et~al.(2022)Grapin, Diouane, Morlier, Bartoli, Lefebvre, Saves, and Bussemaker}]{grapin_constrained_2022}
Grapin, R., Diouane, Y., Morlier, J., Bartoli, N., Lefebvre, T., Saves, P., and Bussemaker, J.~H., \enquote{Regularized Inﬁll Criteria for Multi-objective Bayesian Optimization with Application to Aircraft Design,} \emph{AIAA AVIATION 2022 Forum}, 2022, pp. 1--20.

\bibitem[{Saves et~al.(2024{\natexlab{b}})Saves, Diouane, Bartoli, Lefebvre, and Morlier}]{Mixed_Paul_PLS}
Saves, P., Diouane, Y., Bartoli, N., Lefebvre, T., and Morlier, J., \enquote{High-dimensional mixed-categorical Gaussian processes with application to multidisciplinary design optimization for a green aircraft,} \emph{Structural and Multidisciplinary Optimization}, Vol.~67, 2024{\natexlab{b}}, p.~81.

\bibitem[{Saves et~al.(2023)Saves, Diouane, Bartoli, Lefebvre, and Morlier}]{Mixed_Paul}
Saves, P., Diouane, Y., Bartoli, N., Lefebvre, T., and Morlier, J., \enquote{A mixed-categorical correlation kernel for {Gaussian} process,} \emph{Neurocomputing}, Vol. 550, 2023, p. 126472.

\bibitem[{Priem et~al.(2020)Priem, Bartoli, Diouane, and Sgueglia}]{SEGO-UTB}
Priem, R., Bartoli, N., Diouane, Y., and Sgueglia, A., \enquote{Upper trust bound feasibility criterion for mixed constrained {Bayesian} optimization with application to aircraft design,} \emph{Aerospace Science and Technology}, Vol. 105, 2020, p. 105980.

\bibitem[{Garnett(2023)}]{garnett2023bayesian}
Garnett, R., \emph{Bayesian optimization}, Cambridge University Press, 2023.

\bibitem[{Calandra et~al.(2016)Calandra, Seyfarth, Peters, and Deisenroth}]{calandra2016bayesian}
Calandra, R., Seyfarth, A., Peters, J., and Deisenroth, M.~P., \enquote{Bayesian optimization for learning gaits under uncertainty: An experimental comparison on a dynamic bipedal walker,} \emph{Annals of Mathematics and Artificial Intelligence}, Vol.~76, 2016, pp. 5--23.

\bibitem[{Bouhlel et~al.(2018)Bouhlel, Bartoli, Regis, Otsmane, and Morlier}]{Bouhlel18}
Bouhlel, M.~A., Bartoli, N., Regis, R., Otsmane, A., and Morlier, J., \enquote{Efficient Global Optimization for high-dimensional constrained problems by using the {Kriging} models combined with the Partial Least Squares method,} \emph{Engineering Optimization}, Vol.~50, 2018, pp. 2038--2053.

\bibitem[{Charayron et~al.(2023)Charayron, Lefebvre, Bartoli, and Morlier}]{charayron2023towards}
Charayron, R., Lefebvre, T., Bartoli, N., and Morlier, J., \enquote{Towards a multi-fidelity and multi-objective {Bayesian} optimization efficient algorithm,} \emph{Aerospace Science and Technology}, Vol. 142, 2023, p. 108673.

\bibitem[{Bussemaker et~al.(2024)Bussemaker, Saves, Bartoli, Lefebvre, and Lafage}]{bussemaker2024system}
Bussemaker, J.~H., Saves, P., Bartoli, N., Lefebvre, T., and Lafage, R., \enquote{System Architecture Optimization Strategies: Dealing with Expensive Hierarchical Problems,} \emph{HAL open archive}, 2024.

\bibitem[{Hall{\'e}-Hannan et~al.(2024)Hall{\'e}-Hannan, Audet, Diouane, {Le Digabel}, and Saves}]{halle2024graph}
Hall{\'e}-Hannan, E., Audet, C., Diouane, Y., {Le Digabel}, S., and Saves, P., \enquote{A graph-structured distance for heterogeneous datasets with meta variables,} \emph{ArXiv preprint}, 2024.

\bibitem[{Duvenaud(2014)}]{GP14}
Duvenaud, D., \enquote{Automatic model construction with {Gaussian} processes,} Ph.D. thesis, University of Cambridge, 2014.

\bibitem[{Rossi(2018)}]{MLE}
Rossi, R., \emph{Mathematical statistics: an introduction to likelihood based inference}, John Wiley \& Sons, 2018.

\bibitem[{Li et~al.(2019)Li, Bouhlel, and Martins}]{Li2019}
Li, J., Bouhlel, M.~A., and Martins, J. R. R.~A., \enquote{Data-Based Approach for Fast Airfoil Analysis and Optimization,} \emph{AIAA Journal}, Vol.~57, No.~2, 2019, p. 581–596.
\newblock \doi{10.2514/1.j057129}.

\bibitem[{Cybenko(1989)}]{cybenko1989approximation}
Cybenko, G., \enquote{Approximation by superpositions of a sigmoidal function,} \emph{Mathematics of control, signals and systems}, Vol.~2, No.~4, 1989, pp. 303--314.

\end{thebibliography}

\end{document}